# A Model-Centric Review of Deep Learning for Protein Design

Gregory W. Kyro*†     Tianyin Qiu†     Victor S. Batista†

## Abstract

Deep learning has transformed protein design, enabling accurate structure prediction, sequence optimization, and de novo protein generation. Advances in single-chain protein structure prediction via AlphaFold2, RoseTTAFold, ESMFold, and others have achieved near-experimental accuracy, inspiring successive work extended to biomolecular complexes via AlphaFold Multimer, RoseTTAFold All-Atom, AlphaFold 3, Chai-1, Boltz-1 and others. Generative models such as ProtGPT2, ProteinMPNN, and RFdiffusion have enabled sequence and backbone design beyond natural evolution-based limitations. More recently, joint sequence-structure co-design models, including ESM3, have integrated both modalities into a unified framework, resulting in improved designability. Despite these advances, challenges still exist pertaining to modeling sequence-structure-function relationships and ensuring robust generalization beyond the regions of protein space spanned by the training data. Future advances will likely focus on joint sequence-structure-function co-design frameworks that are able to model the fitness landscape more effectively than models that treat these modalities independently. Current capabilities, coupled with the dizzying rate of progress, suggest that the field will soon enable rapid, rational design of proteins with tailored structures and functions that transcend the limitations imposed by natural evolution. In this review, we discuss the current capabilities of deep learning methods for protein design, focusing on some of the most revolutionary and capable models with respect to their functionality and the applications that they enable, leading up to the current challenges of the field and the optimal path forward.

* Correspondence to gregory.kyro@yale.edu
† Department of Chemistry, Yale University

# Contents

# 1. The Biological Significance of Protein Design

Proteins fundamentally define cellular function,[1] and their structural and dynamic properties dictate their roles in important biochemical events such as catalysis, molecular recognition, and regulation.[2] The ability to design proteins with specific functional properties therefore enables direct modulation of important biochemical processes.[3]

There are already examples of enzymes (i.e., proteins that catalyze chemical reactions by stabilizing transition states) that have been optimized for substrate specificity, enhanced reaction kinetics, and improved stability under diverse conditions.[4] Additionally, cytokines (i.e., small proteins that are secreted by immune cells and bind to specific receptors on target cells to modulate their behavior) and growth factors (i.e., proteins that bind to specific cell surface receptors to stimulate cell growth, proliferation, differentiation, and healing) have been designed for enhanced specificity, thus improving therapeutic outcomes in immune modulation, regenerative medicine, and oncology.[5, 6]

Antibodies have been designed for high-affinity binding with improved selectivity for targeted inhibition of disease-associated proteins.[7, 8] In addition to antibodies, there are multiple instances of synthetic protein binders that have been engineered to bind to a protein target's surface with high affinity to inhibit protein-protein interactions,[9] induce conformational changes that either activate or inhibit function,[10] modulate signaling by blocking natural ligands,[11] and recruit degradation machinery to eliminate disease-causing proteins.[12] These examples, along with many others,[13] demonstrate how rational protein design can enable us to directly modulate key cellular functions and offer important therapeutic solutions.

# 2. Protein Design Objectives

The objectives of protein design can be generally categorized as either refining the function of an existing protein through targeted modifications or constructing a novel protein (i.e., de novo design). Targeted modifications to naturally occurring proteins can improve their stability, specificity, or catalytic efficiency, enabling enhanced performance under different environmental conditions.[14, 15] For example, introducing targeted mutations into enzymes can optimize substrate binding or broaden their operational range to function in non-native conditions.[16] Similarly, altering an existing protein's binding interface can expand its specificity, allowing it to engage with new targets or disrupt previously inaccessible protein-protein interactions.[17]

Beyond optimizing native proteins from improved functionality, rational engineering efforts can repurpose them for entirely new functions;[18] a protein initially evolved for one biological role can be reprogrammed to perform a distinct function through structural adaptation. This approach has been successfully employed in enzyme engineering, where catalytic domains are redesigned to process novel substrates or mediate alternative reaction mechanisms.[19] Likewise, protein binders can be reconfigured to target different ligands, enabling applications in drug discovery and synthetic biology.[20]

De novo protein design provides a more generalizable framework that enables applications requiring functionalities that do not exist in nature.[3] This strategy typically involves identifying structural motifs that support an intended function, constructing a stable backbone structure that accurately presents these motifs, and selecting an amino acid sequence that folds into this structure.[21] This approach has enabled the development of proteins with novel folds,[22] precise molecular recognition properties,[23] and highly stable assemblies.[24] In addition, de novo strategies have proven successful in designing proteins for both protein-protein interactions and small-molecule recognition.[25, 26] In many cases, de novo design is more effective than adapting existing protein structures to achieve a desired function;[21] de novo design offers complete control over the protein structure and sequence, whereas natural proteins are often marginally functional and marginally stable.[3]

## 3. Deep Learning for Protein Design

Protein design is fundamentally constrained by the vast combinatorial space of possible amino acid sequences, as well as the complex, nonlinear relationship between sequence and function.[27] The *fitness landscape* conceptualizes this relationship, defining how sequence variations correspond to functional properties.[28] With a search space of $20^{100}$ (~$1.27 \times 10^{130}$) possible sequences for even moderate-length (100 amino acids) proteins,[29] brute-force exploration is infeasible, necessitating methods that can efficiently model and navigate this landscape.[30]

Deep learning has fundamentally reshaped protein design by enabling data-driven frameworks capable of modeling sequence-structure-function relationships.[31] The abilities to predict protein structure from sequence with high accuracy,[32-40] generate novel sequences optimized for stability and function,[41-43] and design entirely new protein backbones have been realized through a diverse set of deep learning approaches.[44] However, there does not yet exist a single framework that can effectively solve all facets of the protein design problem. Instead, specialized architectures have emerged, each addressing distinct challenges.[45]

Protein structure prediction remains a cornerstone of computational biology, as the function of a protein is largely dictated by its three-dimensional configuration.[46] Deep learning models such as AlphaFold2,[32] RoseTTAFold,[33] and ESMFold[34, 35] have demonstrated remarkable accuracy in predicting single-chain protein structure from an amino acid sequence. These advances, along with others,[47] have inspired the development of models that can predict multi-component biomolecular complex structures such as AlphaFold Multimer,[36] RoseTTAFold All-Atom,[37] AlphaFold 3,[38] Chai-1,[39] and Boltz-1.[40]

In addition to structure prediction, the inverse task of sequence design (i.e., determining amino acid sequences that reliably fold into a target structure) presents another incredibly valuable capability for protein engineering.[47] However, traditional energy-based optimization methods struggle to efficiently explore sequence space, thus motivating the development of learning-based alternatives.[48] ProtGPT2 addresses de novo sequence generation by modeling the statistical distribution of ~50 million protein sequences, capturing sequence-level patterns observed in nature, but lacking explicit structural constraints.[41] ProteinMPNN introduced structure-guided design by learning sequence-structure relationships through a graph-based message-passing framework, significantly improving sequence recovery compared to previous methods.[42] More recently, BindCraft integrates AlphaFold2 gradients into an end-to-end differentiable optimization pipeline, enabling direct refinement of sequences for high-affinity binding to a target protein.[43]

Deep learning has also significantly enhanced our ability to generate novel protein backbones.[49] Traditional approaches rely on fragment-based assembly and energy minimization, which impose constraints on structural diversity.[50] RFdiffusion introduced a deep learning-based alternative which employs a denoising diffusion probabilistic model (DDPM) to iteratively construct protein backbones from noise, enabling de novo fold generation unconstrained by existing templates.[44] This approach has facilitated the design of proteins with novel topologies and functional scaffolds.[44] Subsequent methods such as RFdiffusion All-Atom have extended this approach to generate protein backbones predicted to bind prespecified small molecules.[51]

More recently, there have been advances that aim to jointly model both sequence and structure within a unified generative framework.[52-54] ESM3 introduced a multimodal transformer-based framework that jointly models sequence, structure, and function, enabling sequences to be designed with implicit structural and functional biases.[52] Pinal adapted this concept by incorporating natural language inputs as additional constraints, enabling the specification of functional properties through text-based prompts.[53]

As new models and methods are developed, existing paradigms will be refined, expanded, or replaced. The following sections will examine some of the current state-of-the-art models that are shaping modern protein design, each organized chronologically to reflect the evolution of the field.

# 4. Protein Structure Prediction from Sequence

Predicting protein structure from amino acid sequence has long been a central challenge in computational biology, given the intricate relationship between sequence and three-dimensional configuration.[55] Accurate structure prediction provides insights into protein function, stability, and interactions, enabling applications in drug discovery, protein engineering, and synthetic biology.[56] Traditional approaches rely heavily on homology modeling and physics-based simulations, which, while effective in certain cases, are limited by the availability of homologous structures and computational complexity.[57] Recent advances in deep learning have revolutionized this field, allowing models to capture sequence-structure relationships at unprecedented accuracy and scale.[58] Among these, AlphaFold2 marked a transformative leap, achieving near-experimental accuracy in predicting single-chain protein structure directly from the sequence.[32]

## 4.1 AlphaFold2

AlphaFold2 predicts protein 3D structure from sequence with high accuracy by integrating multiple sequence alignments (MSAs; structured alignments of evolutionarily related sequences that reveal conserved residues and co-evolutionary patterns), structural templates, an iterative attention-based network architecture known as the *Evoformer*, and a structure module which employs an Invariant Point Attention (IPA) mechanism to model spatial relationships.[32]

AlphaFold2 compiles homologous sequences obtained from UniRef90,[59] BFD, Uniclust30,[60] and MGnify[61] databases into MSAs. When available, structural templates from the Protein Data Bank (PDB) are aligned to the query sequence, offering explicit spatial priors to guide the model.[32]

For each residue pair in a given sequence, AlphaFold2 initializes pairwise features by integrating relative positional embeddings with MSA-derived information and structural templates (when available). Positional embeddings encode binned distances between two residues in the sequence, allowing the model to capture dependencies based on sequence proximity. MSA-derived co-evolutionary information is extracted using the outer product mean operation (i.e., for each residue pair, the outer product of their embedding vectors are computed across all homologous sequences in the MSA, which are then averaged across the MSA to form a single co-evolutionary matrix). When available, template-derived structural priors provide explicit pairwise distances and orientations from aligned PDB structures, offering spatial constraints which are particularly valuable for regions with sparse MSA coverage.[32]

The AlphaFold2 architecture is divided into two primary components: the Evoformer and the structure module. The Evoformer network processes MSAs and pairwise residue features through 48 blocks, each comprising MSA attention and pair attention layers. MSA attention applies self-attention both within and between sequences in the MSA, enhancing evolutionary signals by capturing conserved patterns and co-evolutionary relationships. Pair attention applies self-attention across pairwise residue embeddings, enabling the model to refine residue-residue interactions.[32] After processed by the Evoformer, the pair representation and the first row of the MSA representation (single representation) are passed to the structure module, which predicts 3D coordinates.

The structure module consists of eight sequential blocks, each sharing the same weights. To maintain SE(3) equivariance (i.e., ensure that for any rotation $R$ and translation $\mathbf{t}$, applying $(R, \mathbf{t})$ to the input yields the same transformation in the output; formally denoted as $f(Rx + \mathbf{t}) = Rf(x) + \mathbf{t}$), each residue is associated with a local coordinate frame represented by a rotation matrix $R$, which is initialized to the identity matrix (i.e., no initial rotation), and a translation vector $\mathbf{t}$, which is initialized to the origin (i.e., no initial displacement). The single representation, pair representation, and local coordinate frames are passed as input to the IPA module, which applies attention over the single and pair representations while maintaining invariance with respect to the global frame. The IPA module outputs an updated single representation and predicts relative rotations ($\Delta R$) and translations ($\Delta \mathbf{t}$) for each residue, which are iteratively applied to refine the local coordinate frames. The updated frames are then used to predict backbone torsion angles ($\phi, \psi, \omega$) for each residue, from which the backbone atomic positions are computed. Finally, side-chain torsion angles ($\chi$) are predicted using a predefined rotamer library, allowing for precise placement of side-chain atoms in 3D space, completing the full-atom structural prediction.[32]

AlphaFold2 is trained end-to-end, supervised by the frame-aligned point error (FAPE) loss, which serves as the primary structural loss, as well as additional auxiliary losses which account for pairwise residue distance error, MSA reconstruction error, per-residue prediction confidence, side-chain prediction error, and structure violation. FAPE compares the predicted atom positions with the true atom positions under many different alignments of the predicted frame ($R_k$, $\mathbf{t}_k$) to the corresponding true frame. Specifically, for each alignment, the distances between all predicted atom positions and corresponding true atoms positions contribute to the loss, ultimately creating a strong bias for the model to correctly position atoms relative to the local frame of each residue (i.e., correctly position side-chain atoms).[32]

AlphaFold2's ability to achieve near-experimental accuracy redefined the perceived limits of protein structure prediction.[57] The model has not only served as an invaluable tool for structural biology,[62] but has also inspired subsequent developments which are revolutionizing protein design.[33-40, 47]

## 4.2 RoseTTAFold

Unlike AlphaFold2's Evoformer (Section 4.1), which processes MSAs and pairwise residue features, RoseTTAFold uses a three-track architecture that integrates MSAs (1D track), pairwise residue features (2D track), and atomic coordinate representations (3D track). This approach improves structure prediction accuracy, particularly for challenging targets with sparse MSAs.[33]

In addition to single-chain protein structure prediction, RoseTTAFold inherently supports the co-folding of multiple protein sequences by treating them as a single input in the 1D track and simultaneously modeling their inter-chain residue-residue interactions in the 2D and 3D tracks. This strategy allows RoseTTAFold to predict the structure of multi-subunit proteins or protein-protein complexes in a single pass.[33]

## 4.3 ESM-2 & ESMFold

### 4.3.1 ESM-2

ESM-2 introduces a shift in protein sequence modeling by employing a BERT-style encoder-only transformer that is trained to predict masked amino acids directly from raw sequence data from the UniRef dataset,[63] which contains about $10^8$ unique protein sequences. This approach enables the model to implicitly capture evolutionary and structural constraints embedded in the sequence data, effectively offering a much more efficient and scalable alternative to MSAs. In order for the model to be able to generalize to sequences longer than those encountered during training, ESM-2 incorporates Rotary Position Embeddings (RoPE) in place of conventional positional encodings, which apply sinusoidal embeddings directly into the query and key vectors in the self-attention mechanism.[34, 35]

### 4.3.2 ESMFold

ESMFold leverages ESM-2 by predicting 3D protein structures directly from ESM-2-derived sequence embeddings, bypassing the need for MSAs. In addition, ESMFold omits template information entirely, and instead uses ESM-2-derived attention maps, which have been shown to implicitly capture structural constraints. These adjustments result in significantly faster and more scalable inference compared to traditional MSA- and template-dependent models.[34, 35]

Compared to the AlphaFold2 architecture that consists of the Evoformer and a structure module (Section 4.1), ESMFold replaces the Evoformer with a much simpler alternative referred to as the *Folding Block*, while retaining the same structure module. Since ESM-2-derived features are 1D, as opposed to 2D MSAs, the folding block performs standard self-attention over the 1D feature space rather than axial attention, which is used in the Evoformer of AlphaFold2. The remainder of the folding block operations mirror those of the Evoformer, enabling ESMFold to maintain structural prediction performance while improving computational efficiency.[34, 35]

## 5. Multi-Chain and Complex Assembly Structure Prediction

Biological function often arises from biomolecular complexes or macromolecular assemblies involving proteins, nucleic acids, and small molecules.[64] Accurately predicting the structures of these complexes, whether protein-protein, protein-ligand, or protein-nucleic acid complexes, requires modeling that can capture inter-molecular evolutionary signals, interface dynamics, and structural constraints unique to each type of complex.[65]

Single-chain models like AlphaFold2 excel at predicting individual protein structures but lack the ability to model biomolecular complexes.[32] To address this limitation, new architectures have been developed to predict multi-component biomolecular structures with high accuracy.[66]

### 5.1 AlphaFold Multimer

AlphaFold Multimer extends AlphaFold2 (Section 4.1) to predict multi-chain protein complex structures. Specifically, the MSA and pair attention mechanisms within the Evoformer network are adapted to properly capture inter-chain co-evolutionary relationships and interactions, respectively. To address the challenge of chain-order dependence in multi-chain prediction, AlphaFold Multimer employs a permutation-invariant loss function, ensuring that the predicted structure is independent of the input chain order. The model is trained on multi-chain proteins and protein-protein complexes from the PDB.[36]

### 5.2 RoseTTAFold All-Atom

RoseTTAFold All-Atom extends the original RoseTTAFold framework (Section 4.2) to predict the structures of biomolecular assemblies, including proteins, nucleic acids, small molecules, metals, and covalent modifications. For proteins and nucleic acid chains, RoseTTAFold All-Atom retains the representations from RoseTTAFold. For small molecules, metals, and covalent modifications, the model is trained on protein-small molecule, protein-metal, and covalently modified protein complexes sourced from the PDB. On the Continuous Automated Model Evaluation (CAMEO) blind ligand-docking benchmark, RoseTTAFold All-Atom demonstrates superior performance compared to existing methods at the time such as AutoDock Vina.[37]

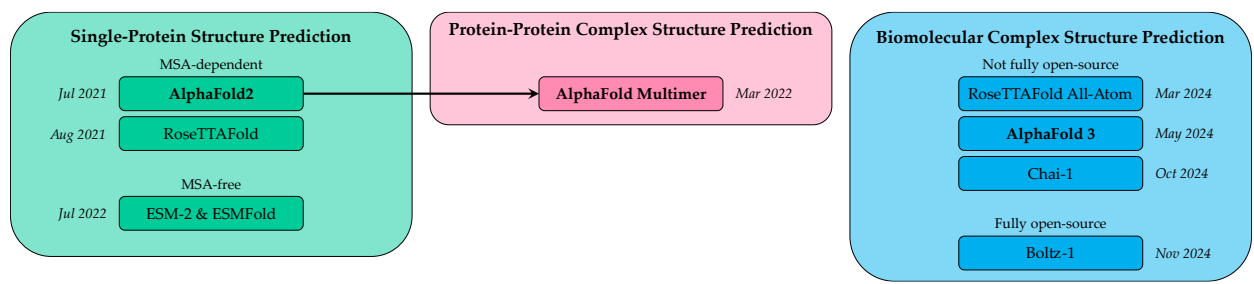

**Figure 1. Evolution of deep learning models for protein structure prediction.** Models are categorized into three major groups: Single-Protein Structure Prediction (left, green), Protein-Protein Complex Structure Prediction (center, pink), and Biomolecular Complex Structure Prediction (right, blue). Single-protein predictors are divided into MSA-dependent (AlphaFold2, RoseTTAFold) and MSA-free (ESMFold) models. AlphaFold Multimer extends AlphaFold2 to predict multi-chain protein structures. Biomolecular complex structure prediction models further generalize to protein-ligand, protein-nucleic acid, and multi-component assemblies.

Beyond structure prediction, RoseTTAFold All-Atom incorporates RFdiffusion All-Atom, a generative diffusion-based model designed to create novel proteins capable of binding specified small molecules. RFdiffusion All-Atom, adapted from RFdiffusion, (Section 6.1) was used to successfully generate binding pockets for ligands including digoxigenin, heme, and bilin which were experimentally validated.[37]

## 5.3 AlphaFold 3

AlphaFold 3 extends the capabilities of AlphaFold2 (Section 4.1) by introducing architectural modifications that enable it to predict the structures of protein-based multi-component biomolecular complexes. Compared to AlphaFold2, AlphaFold 3 replaces the Evoformer with the *Pairformer* network, which retains the sequence and pair representations, but removes the MSA representation. This strategy enables the model to predict structures accurately even when homologous sequences are limited.[38]

Additionally, the deterministic structure module from AlphaFold2 is replaced with a diffusion-based generative model that takes noise-perturbed atomic coordinates as input and iteratively refines them through multiple denoising steps to predict a final 3D structure. SE(3) equivariance is enforced by applying random rotations and translations to the input coordinates during training, ensuring that predictions remain invariant to global orientation and position. The diffusion-based component of the AlphaFold 3 architecture consists of 30 sequential attention blocks; three local attention blocks capture atomic-level interactions, followed by 24 global attention blocks that model long-range residue-residue interactions, concluding with three local attention blocks that refine atomic positions before yielding the final output.[38]

To enable accurate confidence estimation, AlphaFold 3 employs a diffusion *rollout* during training. Although trained on single-step denoising rather than full end-to-end generation, during diffusion rollout, the model simulates the complete diffusion process through 20 iterative denoising steps starting from a noise-perturbed structure. The resulting predicted structure is then compared to the ground truth to yield performance metrics that are used to train a confidence head to predict per-residue confidence scores.[38]

AlphaFold 3's ability to predict the structures of diverse biomolecular complexes and macromolecular assemblies markedly advanced deep learning in structural biology,[38] inspiring subsequent models to emulate its approach.[39, 40]

## 5.4 Chai-1

Chai-1 builds upon the architectural foundation of AlphaFold 3 (Section 5.3) by introducing key modifications that enhance predictive flexibility and accuracy. Specifically, Chai-1 incorporates residue-level embeddings from a 3-billion-parameter protein language model akin to ESM-2 (3B) (Section 4.3.1), which encode contextual sequence patterns and implicit evolutionary signals. These embeddings are combined with sequence and pair representations, akin to those in AlphaFold 3, enabling flexible inference with any combination of MSAs, structural templates, and language model embeddings.[39]

In addition, Chai-1 integrates structural constraints as priors to guide the model during both training and inference. Specifically, pocket constraints enforce binding-site proximity using residue-specific token IDs, chain IDs, and a distance threshold; contact constraints enforce residue-residue proximity using token pairs and a distance threshold; docking constraints encode inter-chain distances as binned pairwise token distances across partitioned chain groups. During training, constraints are included with a 10% probability. When a constraint is not included, a separate learnable mask value is used.[39]

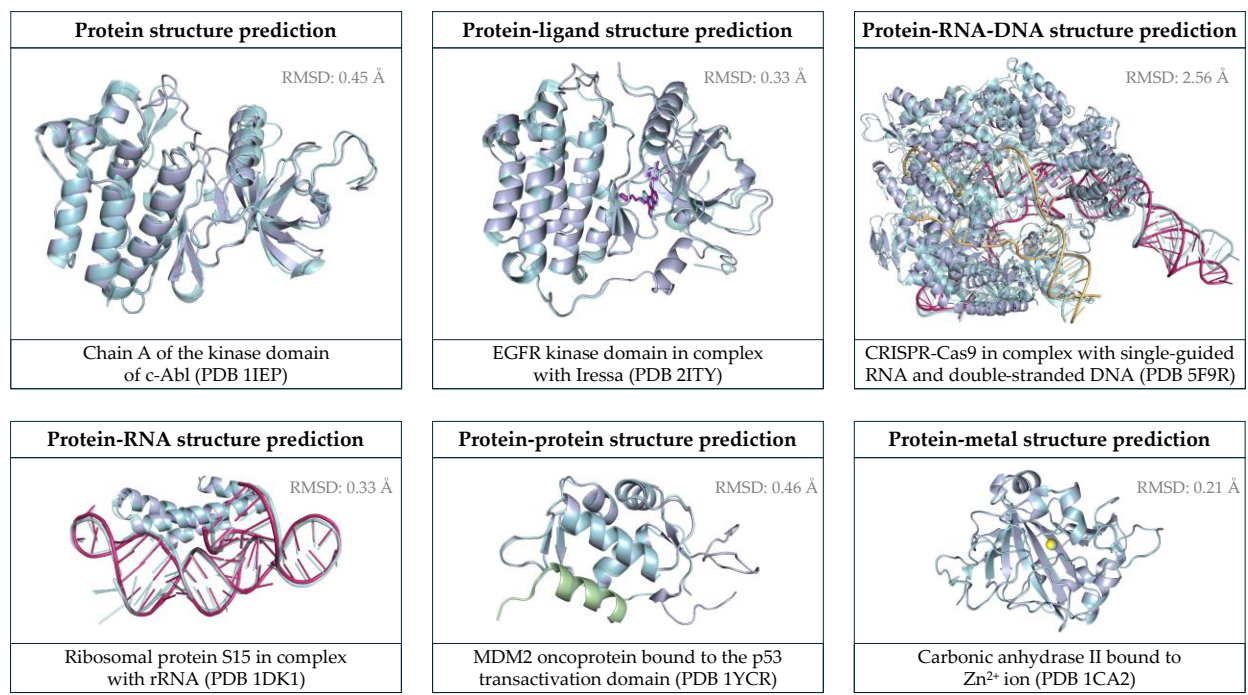

**Figure 2. Biomolecular complex and macromolecular assembly structures predicted by Chai-1.** Predicted structures are compared to the corresponding experimentally determined crystal structures from the Protein Data Bank. Experimentally determined crystal structures are shown in cyan. Chai-1-predicted structures follow the color scheme: proteins (blue), ligands (purple), RNA (pink), DNA (orange), additional protein subunits in protein-protein complexes (green), and metal (yellow). (Top-left): Single-protein structure prediction of chain A of c-Abl kinase (PDB 1IEP),[67] RMSD = 0.45 Å. (Top-center): Protein-ligand complex structure prediction of EGFR kinase bound to Iressa (PDB 2ITY),[68] RMSD = 0.33 Å. (Top-right): Protein-RNA-DNA complex structure prediction of CRISPR-Cas9 bound to single-guide RNA and double-stranded DNA (PDB 5F9R),[69] RMSD = 2.56 Å. (Bottom-left): Protein-RNA complex structure prediction of ribosomal protein S15 bound to rRNA (PDB 1DK1),[70] RMSD = 0.33 Å. (Bottom-middle): Protein-protein complex structure prediction of the MDM2 oncoprotein bound to the p53 tumor suppressor transactivation domain (PDB 1YCR),[71] RMSD = 0.46 Å. (Bottom-right): Protein-metal complex structure prediction of carbonic anhydrase II bound to $Zn^{2+}$ ion (PDB 1CA2),[72] RMSD = 0.21 Å.

By developing a model largely inspired by the AlphaFold 3 architecture while augmenting it with language model embeddings and structural constraints, Chai-1 offers enhanced flexibility and accuracy in predicting complex biomolecular assemblies.[39]

## 5.5 Boltz-1

Boltz-1 represents an open-source recreation of AlphaFold 3 with targeted refinements to the architecture. Specifically, the model modifies the flow of operations in the MSA module (i.e., the module which processes and passes the pair representation to the Pairformer) to improve the interaction between single and pair representations. In addition, Boltz-1 introduces residual connections and reorders operations in the diffusion module's attention layers to improve gradient flow, backpropagation efficiency, and information transfer within layers. Released under the MIT License, Boltz-1 offers greater accessibility for the research community, facilitating further development and adaptation.[40]

# 6. De Novo Structure Generation

De novo protein structure generation aims to design entirely new proteins independently of evolutionary constraints.[49] Traditional approaches rely on fragment-based assembly and physics-based energy minimization, but these methods impose significant constraints on design flexibility and structural diversity.[23] Deep generative models, particularly denoising diffusion probabilistic models (DDPMs), have enabled de novo backbone generation by directly learning structural distributions and iteratively refining random initializations into stable protein folds.[49] This capability enables the creation of novel topologies and functional scaffolds beyond what exists in nature.[73]

## 6.1 RFdiffusion

RFdiffusion, developed by fine-tuning RoseTTAFold (RF), is a DDPM for de novo protein backbone design. The model follows the standard forward (noising)-reverse (denoising) diffusion framework. In the forward process, protein structures derived from the PDB are progressively perturbed over 200 steps by adding Gaussian noise on C$\alpha$ coordinates for translation and using Brownian motion on the manifold of rotation matrices for residue orientations. In the reverse process, a neural network learns to iteratively denoise the protein backbone by predicting the noise applied at each timestep from noisy coordinates and orientations. To ensure trajectory coherence across denoising timesteps, RFdiffusion self-conditions noise predictions at each timestep on previous timestep predictions. During inference, the model starts from completely random residue frames, ultimately generating a protein backbone structure.[44]

Beyond unconditional protein backbone generation from noise, RFdiffusion enables conditional generation with explicit control over design constraints, including functional-motif (e.g., enzyme actives site, viral epitope) scaffolding, topology (e.g., TIM barrel)-conditioned design, and conditioning on hotspot residues for protein binder design. To enable protein binder design, RFdiffusion is fine-tuned on multi-chain proteins and protein-protein complexes from the PDB, learning to generate a binding protein given the hotspot residues of the target protein. There are numerous examples of RFdiffusion demonstrating remarkable success in each of these instances.[44]

Subsequently, RFdiffusion All-Atom (RFdiffusionAA) was developed by fine-tuning RoseTTAFold All-Atom (Section 5.2), and extended the capabilities of RFdiffusion to generate novel protein backbone structures conditioned on specified small molecules to which the generated proteins bind.[37]

# 7. De Novo Sequence Generation and Optimization

Generating novel protein sequences that reliably fold into functional structures represents a fundamental challenge in protein design.[74] Traditional approaches rely on energy-based models and heuristic search methods, which are often computationally expensive and limited in their ability to navigate the vast sequence space.[75] Recent advances in deep learning have introduced generative models capable of designing protein sequences de novo that extend beyond the sequence space produced via natural selection and optimizing sequences for desired properties (e.g., enhanced stability, expression, binding affinity, and/or catalytic activity).[41-43]

## 7.1 ProtGPT2

ProtGPT2 is an autoregressive transformer-based model for de novo protein sequence generation. The model is trained on ~50 million non-annotated protein sequences from the UniRef50 database, enabling it to learn the underlying sequence patterns, residue frequencies, and co-evolutionary relationships that define natural protein families. Unsurprisingly, sequences generated by ProtGPT2 exhibit strong resemblance to natural proteins in terms of amino acid composition, secondary structure motifs, and hydrophobicity patterns. However, because of the stochastic nature of transformer-based generative models, ProtGPT2 can expand existing protein families and explore novel regions of sequence space that are not spanned by the training data.[41]

## 7.2 ProteinMPNN

ProteinMPNN is a graph-based message-passing neural network (MPNN) for protein sequence design, containing a backbone structure encoder and sequence decoder for predicting amino acid sequences that fold into a specified backbone structure. Unlike energy-based methods like Rosetta, which rely on computationally intensive Monte Carlo optimization to design low-energy sequences for fixed backbones,[76] ProteinMPNN predicts optimal sequences using learned sequence-structure relationships, resulting in significantly improvements in sequence recovery (52.4% vs. 32.9% achieved by Rosetta) and design accuracy.[42]

The model represents protein backbone structures as graphs, where nodes correspond to C$\alpha$ atoms and edges exist between each node and its 30 closest neighbors as determined by Euclidean distance in 3D space. Edges encode pairwise geometric features including inter-residue C$\alpha$–C$\alpha$ distances, intra-residue pairwise distances between N, C$\alpha$, C, O, and virtual C$\beta$ atoms, backbone dihedral angles, and local frame orientations. This graph is processed by the encoder, which is an MPNN that updates both the nodes and edges. The encoded features are then processed by an order-agnostic autoregressive decoder that predicts a probability distribution over amino acids at each position in the sequence. Order-agnostic decoding is achieved by randomizing the decoding order.[42]

ProteinMPNN is trained on 19,700 high-resolution single-chain structures from the PDB. The objective of the model during training is to minimize per-residue cross-entropy loss between predicted and native sequences given a protein backbone structure as input. To enhance generalization, especially to predicted structures, the model perturbs backbone coordinates with Gaussian noise (standard deviation of 0.02 Å) during training.[42]

ProteinMPNN was applied to successfully rescue failed Rosetta designs for protein binders, and produce stable, expressible proteins that were validated through X-ray crystallography and cryo-EM.[42]

Although originally developed for fixed-backbone sequence design, ProteinMPNN was repurposed to optimize protein sequences for improved expression, stability, and function. This was achieved by fixing residues critical to function (e.g., ligand-binding sites), leveraging evolutionary conservation to maintain structurally or functionally important positions, using ProteinMPNN to generate sequence variants predicted to fold into the specified backbone structure, and filtering the designed sequences using AlphaFold2 to ensure high-confidence folding and structural agreement with the input backbone structure. This approach was successfully applied to myoglobin to enhance stability and function at elevated temperatures, as well as to TEV protease to enhance stability and catalytic efficiency.[77]

**Table 1.** Selected deep learning models for protein design, categorized by primary function and sorted by date of paper release in ascending chronological order.

| Model | Paper Date | Primary Function | Key Features and Capabilities |
| --- | --- | --- | --- |
| AlphaFold2 | Jul 15, 2021 | Single-chain protein structure prediction | MSA-based transformer with Evoformer and IPA; high-accuracy single-chain protein structure prediction |
| RoseTTAFold | Aug 19, 2021 | Single-chain protein structure prediction | Three-track network integrating MSA, pairwise, and 3D structural features for single-chain protein structure prediction |
| ESM-2 | Jul 21, 2022 | Sequence representation | MSA-free, BERT-style transformer trained on large-scale protein sequences for learned embeddings |
| ESMFold | Jul 21, 2022 | Single-chain protein structure prediction | MSA-free structure predictor using ESM-2 embeddings with a simplified folding module |
| AlphaFold Multimer | Mar 10, 2022 | Multi-chain protein complex prediction | Extension of AlphaFold2 with multi-chain MSAs and permutation-invariant loss for complex modeling |
| RoseTTAFold All-Atom | Mar 07, 2024 | Biomolecular complex prediction | Unified framework for biomolecular complexes of proteins, nucleic acids, small molecules, and metals |
| AlphaFold 3 | May 08, 2024 | Biomolecular complex prediction | MSA-free diffusion model with Pairformer for multi-component biomolecular complexes |
| Chai-1 | Oct 11, 2024 | Biomolecular complex prediction | AlphaFold 3 variant integrating structural constraints and protein language model embeddings |
| Boltz-1 | Nov 20, 2024 | Biomolecular complex prediction | Open-source AlphaFold 3 reimplementation with optimized MSA and diffusion modules |
| RFdiffusion | Jul 11, 2023 | De novo backbone structure generation | Diffusion model for both unconditional and conditional de novo backbone generation |
| RFdiffusion All-Atom | Mar 07, 2024 | Small molecule-conditioned backbone structure generation | Successor to RFdiffusion that enables generation of protein backbones conditioned on small molecules to which the generated proteins bind |
| ProtGPT2 | Jul 27, 2022 | De novo sequence generation | Autoregressive transformer trained on UniRef50 for de novo sequence generation |
| ProteinMPNN | Sep 15, 2022 | Structure-guided sequence optimization | Graph-based MPNN with order-agnostic decoding for sequence design for fixed backbones |
| FAMPNN | Feb 17, 2025 | Joint sequence-sidechain optimization | Simultaneously predicts sequence identity via masked language modeling and sidechain coordinates via denoising of perturbed atom positions |
| ESM3 | Jul 02, 2024 | Joint sequence-structure-function co-design | Multimodal transformer integrating sequence, structure, and function constraints for protein design |
| Pinal | Aug 02, 2024 | Joint sequence-structure-function co-design | Text-to-structure-to-sequence pipeline for text-guided design of protein sequences from natural language descriptions |

The model is trained using a generative masked language modeling objective over sequence, structure, and function tokens, where a random subset of tokens from each track is masked and the model is optimized to reconstruct them from the unmasked context, forcing the model to learn interdependencies between the modalities. The model applies supervision across various masking rates rather than a single fixed rate, ensuring robustness to different levels of missing information. During generation, masked tokens are iteratively sampled in any order until all masked positions are predicted, allowing ESM3 to generate proteins from any combination of sequence, structure, or function inputs.[52]

Recent work builds upon ProteinMPNN by jointly modeling sequence identity and sidechain structure within a combined loss objective. In particular, Full-Atom MPNN (FAMPNN) demonstrates that this strategy improves both sequence recovery and sidechain atomic coordinate prediction accuracy.[78]

## 7.3 BindCraft

BindCraft is an automated pipeline for de novo protein binder design that co-optimizes binder backbone and sequence by backpropagating a composite loss through the fixed weights of AlphaFold Multimer to refine the sequence for high-confidence target interactions while subsequent AlphaFold2 monomer predictions ensure standalone structural stability.[43]

The BindCraft pipeline begins by randomly initializing a binder sequence and pairing it with the target protein sequence. The target-binder complex sequence is then processed through AlphaFold Multimer, which predicts their joint 3D co-folded structure. A multi-objective loss function then refines the binder sequence, incorporating per-residue confidence scores for binder structural quality, predicted alignment error for confidence in the binder's internal structure and its interface with the target, interface contact loss for strong binder-target binding, and binder helicity and compactness constraints for a stable fold. Gradients of this loss are backpropagated to iteratively update the sequence for both target binding and standalone stability.[43]

Sequence optimization via backpropagation through AlphaFold Multimer generates binders with high in silico confidence but often poor in vitro expression due to unrealistic amino acid distributions. BindCraft addresses this by incorporating ProteinMPNN to re-optimize the sequences for enhanced expression, solubility, and stability while keeping the binding interface fixed. The ProteinMPNN-refined sequences are then re-predicted by AlphaFold2. The predicted designs are then filtered based on AlphaFold2 confidence metrics, as well as Rosetta physics-based scoring metrics to eliminate designs that are physically improbable. Remarkably, BindCraft demonstrated a 49.5% average success rate across 10 diverse protein targets including challenging targets such as CRISPR-Cas9, significantly outperforming prior binder design approaches.[43]

## 8. Joint Sequence-Structure Co-Design

Joint sequence-structure co-design has emerged as a powerful approach to protein design, enabling the simultaneous modeling of amino acid sequences and their corresponding three-dimensional structures within a unified generative framework.[79] Traditional methods typically model either sequence or structure individually, either designing sequences for predefined backbones or predicting structures from sequences, limiting the ability to model their interdependence.[80] Recent deep learning models have enabled the co-design of sequence and structure, capturing their intrinsic dependencies to improve stability, functionality, and designability.[81, 82]

### 8.1 ESM3

ESM3 is a multimodal encoder-decoder transformer, distinct from the encoder-only ESM-2 model (Section 4.3.1), that jointly models protein sequence, structure, and function. In the ESM3 architecture, the encoder maps input sequence, structure, and function tokens to high-dimensional latent embeddings, and the decoder generates novel sequences by attending to these encoded representations. Sequence is tokenized as discrete amino acid residues; structure is encoded by a discrete autoencoder that compresses 3D atomic coordinates into per-residue tokens, which are further refined via a geometric attention mechanism that uses residue-local reference frames to enforce spatial invariance and capture backbone geometry; function is represented as embedded keyword tokens from InterPro and Gene Ontology annotations.[83, 84] These three representations are fused within the model's shared latent space, and then processed via stacked transformer blocks which enable the model to learn complex dependencies between sequence, structure, and function.[52]

### 8.2 Pinal

Pinal introduces a two-stage framework for text-guided protein design, leveraging deep learning to translate natural language descriptions into protein structures and subsequently into amino acid sequences. The model is trained on ~1.7 billion protein (structure or sequence)-text (natural language description) pairs.[53]

The first component of the framework, *T2struct*, learns a mapping from natural language descriptions to structures. Structures are encoded as discrete tokens via vector quantization, where local structural motifs are clustered into a fixed vocabulary and mapped to learnable embeddings. The second component of the framework, *SaProt-T*, learns a mapping from T2Struct-generated backbone structures and their corresponding natural language descriptions to amino acid sequences. Specifically, Pinal explicitly models the joint probability distribution:

$$P(\text{sequence}|\text{text}) = P(\text{structure}|\text{text})P(\text{sequence}|\text{structure}, \text{text})$$

where $P(\text{structure}|\text{text})$ is achieved by T2struct, and $P(\text{sequence}|\text{structure}, \text{text})$ is achieved by SaProt-T. Ultimately, the framework learns to generate amino acid sequences from natural language descriptions.[53]

# 9. Outlook and Future Directions

The advent of deep learning has catalyzed a paradigm shift in protein design, supplanting traditional physics-based methodologies with data-driven, end-to-end differentiable frameworks.[85] This transformation has provided exceptional capabilities for high-accuracy single-chain protein structure prediction, as exemplified by AlphaFold2,[32] RoseTTAFold,[33] and ESMFold,[34, 35] and others,[86] which have achieved near-experimental accuracy. The extension of these capabilities to multi-chain systems and biomolecular assemblies, through models such as AlphaFold Multimer,[36] RoseTTAFold All-Atom,[37] AlphaFold 3,[38] and others,[39, 40] has significantly broadened their utility, with many examples of demonstrated success in drug discovery and synthetic biology.[87-128]

Generative models such as RFdiffusion,[44] ProteinMPNN,[42] and others,[41] have further revolutionized the field by enabling the de novo design of proteins with novel folds and optimized functional properties,[129] with many successful applications already demonstrated.[42, 130-174] Moreover, the emergence of joint sequence-structure co-design frameworks, such as ESM3,[52] has markedly improved designability by learning a probability distribution over both sequences and structures, thus capturing their interdependence.

Despite these remarkable advances, key challenges remain. For instance, accurately modeling the intricate relationships between sequence, structure, and function remains elusive, particularly in the context of generalization beyond the regions of protein space spanned by the training data.[30, 175, 176] In addition, issues such as limited diversity and comprehensiveness of available data, as well as limited realistic representations of complete protein conformation ensembles limit potential progress.[177] Addressing these challenges will require concerted efforts in data curation.

Looking ahead, future breakthroughs in protein design will likely be driven by advancements in joint sequence-structure-function co-design frameworks that are able to model the fitness landscape more effectively than models that treat these modalities independently. As these technologies mature, deep learning will increasingly enable the rapid, rational design of proteins with tailored structures and functions, transcending the limitations imposed by natural evolution.

## 10. Code Repositories and Interactive Notebooks

To facilitate practical application, we provide a curated list of repositories hosted on GitHub or Hugging Face, along with Google Colab notebooks, for the deep learning models reviewed in this article.

### AlphaFold2 & AlphaFold Multimer
- **GitHub:** github.com/google-deepmind/alphafold
- **Colab:** github/sokrypton/ColabFold/blob/main/AlphaFold2.ipynb

### RoseTTAFold
- **GitHub:** github.com/RosettaCommons/RoseTTAFold
- **Colab:** github/sokrypton/ColabFold/blob/main/RoseTTAFold.ipynb

### ESMFold
- **GitHub:** github.com/facebookresearch/esm
- **Colab:** github/sokrypton/ColabFold/blob/main/ESMFold.ipynb

### RoseTTAFold All-Atom
- **GitHub:** github.com/baker-laboratory/RoseTTAFold-All-Atom

### AlphaFold 3
- **GitHub:** github.com/google-deepmind/alphafold3

### Chai-1
- **GitHub:** github.com/chaidiscovery/chai-lab

### Boltz-1
- **GitHub:** github.com/jwohlwend/boltz
- **Colab:** github/sokrypton/ColabFold/blob/main/Boltz1.ipynb

### RFdiffusion
- **GitHub:** github.com/RosettaCommons/RFdiffusion
- **Colab:** github/sokrypton/ColabDesign/blob/main/rf/examples/diffusion.ipynb

### RFdiffusion All-Atom
- **GitHub:** github.com/baker-laboratory/rf_diffusion_all_atom
- **Colab:** github/Graylab/DL4Proteins-notebooks/blob/main/notebooks/WS10_RFDiffusion_AllAtom.ipynb

### ProtGPT2
- **Hugging Face:** huggingface.co/nferruz/ProtGPT2
- **Colab:** colab.research.google.com/drive/14opLMXoPd2y_Hxiu7ZwmeUAUvlB0qu64?usp=sharing

### ProteinMPNN
- **GitHub:** github.com/dauparas/ProteinMPNN
- **Colab:** github/dauparas/ProteinMPNN/blob/main/colab_notebooks/quickdemo.ipynb

### FAMPNN
- **GitHub:** github.com/richardshuai/fampnn

### BindCraft
- **GitHub:** github.com/martinpacesa/BindCraft
- **Colab:** github/martinpacesa/BindCraft/blob/main/notebooks/BindCraft.ipynb

### ESM3
- **GitHub:** github.com/evolutionaryscale/esm
- **Colab:** github.com/evolutionaryscale/esm/tree/main/tools

### Pinal
- **GitHub:** github.com/westlake-repl/Denovo-Pinal

## Acknowledgments

We acknowledge financial support from the National Science Foundation Graduate Research Fellowship under Grant DGE-2139841 [GWK], from the National Science Foundation Engines Development Award: Advancing Quantum Technologies (CT) under Award Number 2302908 [VSB], and from the CCI Phase I: National Science Foundation Center for Quantum Dynamics on Modular Quantum Devices (CQD-MQD) under Award Number 2124511 [VSB].

## Author Information

**Corresponding Author:**
- Gregory W. Kyro
  - Email: gregory.kyro@yale.edu

**Other Authors:**
- Tianyin Qiu
  - Email: tianyin.qiu@yale.edu
- Victor S. Batista
  - Email: victor.batista@yale.edu

**Present Address:** Department of Chemistry, Yale University, New Haven, CT 06511-8499

**Author Contributions:** GWK wrote the paper; TQ and VSB provided feedback on the paper.

**Funding Sources:**
- National Science Foundation Graduate Research Fellowship: Grant DGE-2139841
- National Science Foundation Engines Development Award – Advancing Quantum Technologies (CT): Award Number 2302908
- CCI Phase I – National Science Foundation Center for Quantum Dynamics on Modular Quantum Devices (CQD-MQD): Award Number 2124511

(86) Chen, L.; Li, Q.; Nasif, K. F. A.; Xie, Y.; Deng, B.; Niu, S.; Pouriyeh, S.; Dai, Z.; Chen, J.; Xie, C. Y. AI-Driven Deep Learning Techniques in Protein Structure Prediction. *International Journal of Molecular Sciences* **2024**, *25* (15), 8426.

(87) Omidi, A.; Møller, M. H.; Malhis, N.; Bui, J. M.; Gsponer, J. AlphaFold-Multimer accurately captures interactions and dynamics of intrinsically disordered protein regions. *Proceedings of the National Academy of Sciences* **2024**, *121* (44), e2406407121.

(88) Pogozheva, I. D.; Cherepanov, S.; Park, S.-J.; Raghavan, M.; Im, W.; Lomize, A. L. Structural modeling of cytokine-receptor-JAK2 signaling complexes using AlphaFold multimer. *Journal of chemical information and modeling* **2023**, *63* (18), 5874-5895.

(89) Bellinzona, G.; Sassera, D.; Bonvin, A. M. Accelerating protein–protein interaction screens with reduced AlphaFold-Multimer sampling. *Bioinformatics Advances* **2024**, *4* (1), vbae153.

(90) Homma, F.; Lyu, J.; van der Hoorn, R. A. Using AlphaFold Multimer to discover interkingdom protein–protein interactions. *The Plant Journal* **2024**, *120* (1), 19-28.

(91) Desai, D.; Kantliwala, S. V.; Vybhavi, J.; Ravi, R.; Patel, H.; Patel, J. Review of AlphaFold 3: Transformative Advances in Drug Design and Therapeutics. *Cureus* **2024**, *16* (7), e63646. DOI: 10.7759/cureus.63646 From NLM.

(92) Nussinov, R.; Zhang, M.; Liu, Y.; Jang, H. AlphaFold, allosteric, and orthosteric drug discovery: Ways forward. *Drug Discov Today* **2023**, *28* (6), 103551. DOI: 10.1016/j.drudis.2023.103551 From NLM.

(93) Kyro, G. W.; Smaldone, A. M.; Shee, Y.; Xu, C.; Batista, V. S. T-ALPHA: A Hierarchical Transformer-Based Deep Neural Network for Protein–Ligand Binding Affinity Prediction with Uncertainty-Aware Self-Learning for Protein-Specific Alignment. *Journal of Chemical Information and Modeling* **2025**. DOI: 10.1021/acs.jcim.4c02332.

(94) Anusha; Zhang, Z.; Li, J.; Zuo, H.; Mao, C. AlphaFold 3–Aided Design of DNA Motifs To Assemble into Triangles. *Journal of the American Chemical Society* **2024**, *146* (37), 25422-25425.

(95) Bohrer, R. A.; Bargmann, B. AlphaFold 3, AI, Antibody Patents, the Future of Broad Pharmaceutical Patent Claims and Drug Development. *AI, Antibody Patents, the Future of Broad Pharmaceutical Patent Claims and Drug Development (October 10, 2024)* **2024**.

(96) Gadde, N.; Dodamani, S.; Altaf, R.; Kumar, S. Leveraging AlphaFold 3 for Structural Modeling of Neurological Disorder-Associated Proteins: A Pathway to Precision Medicine. *bioRxiv* **2024**, 2024.2011. 2018.624211.

(97) Zonta, F.; Pantano, S. From sequence to mechanobiology? Promises and challenges for AlphaFold 3. *Mechanobiology in Medicine* **2024**, *2* (3), 100083.

(98) Sun, H.; Li, C.; Pu, Z.; Lu, Y.; Wu, Z.; Zhou, L.; Lin, H.; Wang, Y.; Zi, T.; Mou, L. Single-cell RNA sequencing and AlphaFold 3 insights into cytokine signaling and its role in uveal melanoma. *Frontiers in Immunology* **2025**, *15*, 1458041.28